# Automatic Large-Scale Data Acquisition via Crowdsourcing for Crosswalk Classification: A Deep Learning Approach

Rodrigo F. Berriel[a], Franco Schmidt Rossi[a], Alberto F. de Souza[a], Thiago Oliveira-Santos[a]

[a]*Laboratório de Computação de Alto Desempenho, Departamento de Informática, Universidade Federal do Espírito Santo, Brazil*

## Abstract

Correctly identifying crosswalks is an essential task for the driving activity and mobility autonomy. Many crosswalk classification, detection and localization systems have been proposed in the literature over the years. These systems use different perspectives to tackle the crosswalk classification problem: satellite imagery, cockpit view (from the top of a car or behind the windshield), and pedestrian perspective. Most of the works in the literature are designed and evaluated using small and local datasets, i.e. datasets that present low diversity. Scaling to large datasets imposes a challenge for the annotation procedure. Moreover, there is still need for cross-database experiments in the literature because it is usually hard to collect the data in the same place and conditions of the final application. In this paper, we present a crosswalk classification system based on deep learning. For that, crowdsourcing platforms, such as OpenStreetMap and Google Street View, are exploited to enable automatic training via automatic acquisition and annotation of a large-scale database. Additionally, this work proposes a comparison study of models trained using fully-automatic data acquisition and annotation against models that were partially annotated. Cross-database experiments were also included in the experimentation to show that the proposed methods enable use with real world applications. Our results show that the model trained on the fully-automatic database achieved high overall accuracy (94.12%), and that a statistically significant improvement (to 96.30%) can be achieved by manually annotating a specific part of the database. Finally, the results of the cross-database experiments show that both models are robust to the many variations of image and scenarios, presenting a consistent behavior.

## 1. Introduction

Correctly identifying crosswalks is an essential task for the driving activity and mobility autonomy. In this context, the crosswalk classification is an important topic of research. Moreover, walking in the sideways and crossing streets may seem like easy and effortless activities. However, people with disabilities, specially the almost 285 million visually impaired people worldwide [1], have several challenges regarding mobility autonomy. The importance of this task is directly connected to the danger of harming pedestrians. Despite this need, there are few crosswalk-related data publicly available on crosswalks worldwide (both their locations and datasets). In addition, crosswalks are often aging, i.e. the painting is fading away; occluded by vehicle and pedestrians; darkened by strong shadows and many other factors. These factors turn the crosswalk classification problem into a particularly challenging task. These factors are specially challenging in developing countries, such as Brazil, where maintenance of the roads and paintings is worse than in developed countries.

Crosswalks can be detected in several perspectives: top-view and satellite imagery; from behind the windshield and from the top of the car (cockpit view); and from the pedestrian perspective. Each perspective has a different purpose, i.e. target different potential applications. Although the general problem of crosswalk classification can be tackled in all these perspectives, the scope of this paper is restricted to the perspective of the cockpit view only.

Many crosswalk classification systems have been proposed in the literature over the years [2, 3, 4, 5, 6, 7, 8]. There are several different applications, ranging from helping the visually impaired people [2, 3, 5], road management [4, 6], advanced driver assistance systems [7], and many others. Many of them use image processing techniques to extract features from the images and apply machine learning algorithms to perform the classification task.

Although the problem has been widely studied in the literature over the years, there are still limitations. Most of these works were designed and evaluated based on small databases. In addition, most of the database are local, i.e. they are limited to a specific neighborhood, city or small country. Furthermore, most of these studies are performed in developed countries. Developing countries, such as Brazil, present a wide range of challenges that may not be encountered in those small datasets of developed countries. Beyond the limitation in the scope of these studies, many of them manually labeled those small datasets in order to develop and evaluate their models. This manual process is hardly scalable, forbidding the application on large-scale databases.

In this paper, we present a crosswalk classification system

*Email address:* `rfberriel@inf.ufes.br` (Rodrigo F. Berriel)

based on deep learning. The system exploits crowdsourcing platforms in order to automatically train a Convolutional Neural Network to perform crosswalk classification in a developing country: Brazil. The proposed system can be used to classify crosswalks in real-time, such as in advanced driver assistance systems and autonomous vehicles. Additionally, this work proposes a comparison study of models trained using the fully-automatic data acquisition and annotation against models that were partially annotated manually. Partial manual annotations use negative automatic labeled samples plus positive manually annotated samples, i.e., it is a refinement of the automatic annotations. Moreover, focus is given on evaluating the system on data from real-world application, i.e. comparing to images from others sources (such as different cameras) and not only to images coming from the crowdsourcing platform. The results show that the models can accurately classify crosswalks using the automatic training (average overall accuracy of $94.12 \pm 0.21\%$) and that manually annotating a specific part of the data can boost the performance of the system (average overall accuracy of $96.30 \pm 0.14\%$). Additionally, despite of being developed and evaluated with focus in Brazil, these models presented robustness when applied to different image sources, being scalable to use in other countries. Finally, the results indicate that the proposed system is able to automatically train models to be used on real-world applications.

## 2. Related Works

There are several approaches to the crosswalk classification problem in the literature. In the perspective of helping people with disabilities, specially the visually impaired people, using cameras available on the phones is a strategy commonly used. A prototype for a cell phone application was developed by Ivanchenko et al. [2] to help the visually impaired people. When approaching an intersection, the application detects crosswalks in the images (if any crosswalk is visible) and helps the user to align in the direction of the crosswalk to safely cross the streets. The feature extraction used in the application is based on simple image processing (edge detection using derivative filters). A limitation of this work is related to the scope of the dataset used in the evaluation: only 90 images with only 30 of them containing crosswalks. In addition, there are some thresholds that may require a tuning phase. Poggi et al. [9] also investigated this problem from the perspective of a pedestrian. The authors developed a wearable mobility aid system that captures RGB-D images. The RGB image is filtered based on the depth (point cloud) information to remove noise. Their system predicts 4 classes of crosswalks (considering different angles) and one class for the other (negative) cases. The authors captured about 2500 images to train a Convolutional Neural Network (approximately 500 for each class), and evaluated on a validation set of 10,165 frames. They reported an accuracy of 88.97%, and 91.59% after the head and pose refinement. As the authors neither described the process used while capturing the dataset nor the location where it was captured nor the diversity of the validation set, it is difficult to assess the quality of the results. Wang et al. [3] also used RGB-D images to propose a wayfinding and navigation aid to improve autonomy of the visually impaired people. For the stair and crosswalk detection and recognition part, their system uses image processing on the RGB image and extract depth feature that are both used on a SVM classifier to distinguish stairs from crosswalks. Like many of the systems in the literature, their dataset is small (228 images with only 30 crosswalks) and local. The authors report an accuracy of 78.90% for the crosswalk class. Considering only the crosswalks and the negative class, the overall accuracy of the system is of 90.98%.

In the aerial (top-view) perspective, Riveiro et al. [4] developed an algorithm for automatic detection of zebra crossings from mobile LiDAR data. The algorithm comprises image segmentation followed by several image processing techniques. From a total of 30 crosswalks, 25 crosswalks were correctly classified (83.33%). The authors analyzed the missed crosswalks and the failures came from painting deterioration, very common on developing countries; and occlusion produced by other vehicles, also very common in traffic. Ghilardi et al. [5] proposed a model that used satellite imagery to perform crosswalk detection and localization in order to help the visually impaired to approach intersections. The best model uses a Local Binary Patter (LBP) feature extraction method and a SVM classifier. The database comprises only 900 image patches (600 for the training and 300 for the testing) from 4 different cities. Ahmetovic et al. [10] presented a method combining two perspectives: aerial (satellite) and street view. Their system is targeted to blind travelers and comprises two steps. Initially, crosswalks are searched in the satellite imagery. Subsequently, only the crosswalks candidates detected for the satellite processing are further validated. Finally, Google Street View panoramas are iteratively acquired and the image is either confirmed as containing a crosswalk or rejected. The Google Street View method was evaluated in only 406 portions of Street View images, which comprises a limitation to the results. Moreover, only images from a single city (San Francisco, CA) were used, and the authors tuned the parameters of the system using images from the same region.

In the context of exploiting crowdsourcing platforms to resolve related problems, Hara and Froehlich [11] used crowdsourcing, computer vision and machine learning to visualize and characterize accessibility at scale. Basically, they apply computer vision and machine learning algorithms and provide a user interface to annotate Google Street View panorama imagery. Automatic curb ramp detection was performed using computer vision, and a machine learning algorithm was used as workflow controller.

As can be seen in the related works, there is still need for studies using large amounts of data and greater diversity, i.e. large-scale database. Increasing the scale of the database introduces other challenges, such as the annotation of such large amount of data. Most of the works hereby reviewed used manual annotations for both training and testing. Manually labeling data at scale is a tedious and prone to error task. Therefore, there is still need for method that not only automatically acquire, but also automatically annotate such large-scale databases.

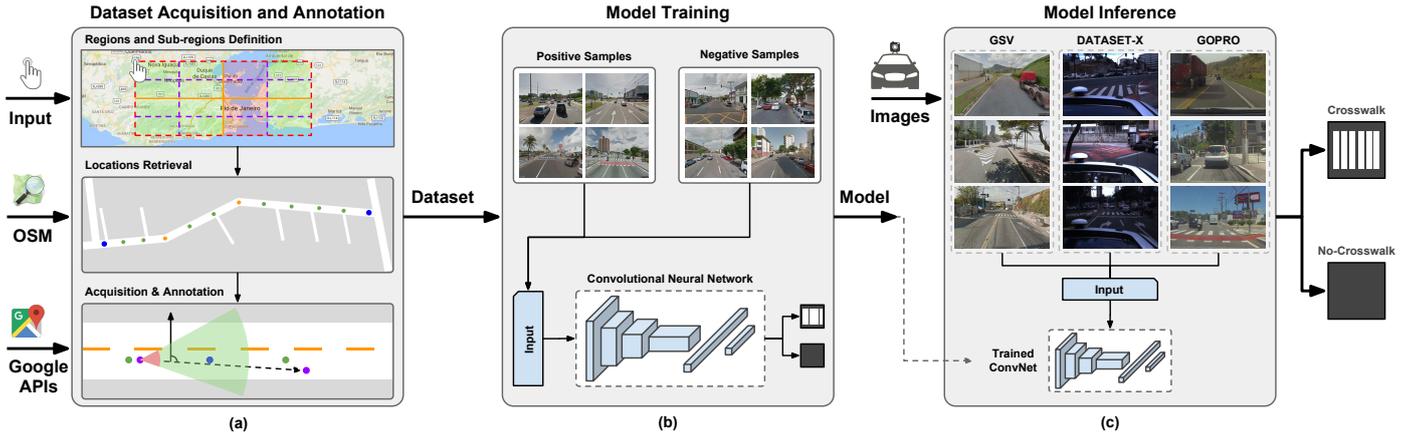

Figure 1: Overview of the proposed system. Firstly, in the dataset acquisition and annotation (a), the system receives the user input to define the regions of interest, then retrieve the locations (both positive and negative) of interest using the OpenStreetMap (OSM) and Google Maps Directions API, and, finally, automatically acquires (using Google Street View Image API) and annotates the samples outputting a dataset. Secondly, in the model training (b), this dataset is used to train a convolutional neural network to predict whether or not there is a crosswalk in a given image, outputting a trained model. Finally, during the model inference (c), images from different sources (such as different cameras) are given to the model and it yields the probability of a given image to contain crosswalks or not.

Moreover, most of these works use an evaluation protocol that allows testing images to be similar to the images seen by the models during training/validation, i.e. images from the same region. In this context, besides improving the evaluation protocol, there is still need for cross-database studies, i.e. training with images from a source and evaluating on other databases from different sources. The proposed work tackles all these needs: large-scale database, automatic acquisition and annotation of data, and cross-database experiments.

## 3. Crosswalk Detection System

The system comprises two main parts: automatic data acquisition and CNN model training/inference. The automatic data acquisition can be performed having just a selection of the area to be acquired. The data is treated, downloaded, and automatically annotated. Alternatively, the data can be partially annotated if desired. With the data, the CNN model can be trained and later used for inference of a given image. The output of the system is a label indicating the presence or not of the crosswalk in a given image. An overview of the proposed system can be seen in the Figure 1.

### 3.1. Dataset Acquisition and Annotation

Acquisition of a large dataset is primordial for automatically training a Convolutional Neural Network. Therefore, the first step of the proposed system is the automatic acquisition and annotation of a large-scale database of images using the APIs of the Google Street View, Google Directions and OpenStreetMap. This acquisition process starts with the user defining the regions of interest. Subsequently, the system automatically downloads the data (positive and negative samples) to identify the presence of crosswalks. Finally, the actual imagery is downloaded and automatically annotated. Alternatively, the user can manually annotate part of the data to correct errors of the automatic system and increase accuracy of the final crosswalk classification.

### 3.1.1. Regions and Sub-regions Definition

The only user interaction necessary is the definition of the region of interest. A region of interest is defined as a rectangular region in world-coordinate system in which the user aims acquiring images (both positive and negative samples). The user can provide one or multiple regions of interest. Each region is defined by two points and must be defined regarding world coordinates. The points required are the bottom-left and the top-right of the region.

For each region of interest, the proposed system retrieves crosswalk locations using the OpenStreetMap[1] (OSM) via Overpass API[2]. One of the limitations imposed by the Overpass API is the size of each of these regions. The maximum allowed size, in any dimension (width or height), of a requested region to the Overpass API is 1/4 degrees. Whenever one of the dimensions of a region is greater than 1/4 degrees, the request is likely to be refused. Therefore, regions greater than this limit are automatically divided into multiple sub-regions. This process is illustrated in the Figure 2, where the region in black solid line has both dimensions greater than the limitation and the orange dashed line are the boundaries of the sub-regions A, B, C and D. At this point, each sub-region dimension is smaller than 1/4 degrees and all sub-regions have the same dimension, i.e. the same height and the same width, although regions may be rectangular (i.e. width and height may be different from one another). When the sub-regions are determined, i.e all sub-regions are complying with the Overpass API constraint, the requests for crosswalk locations are dispatched. OpenStreetMap has several tags referring to pedestrian crossings. In this work, the tag `highway=crossing` was used. This tag has almost two times more nodes associated with it than one of the "useful combinations" (`crossings=*`). Moreover, this tag is the recommended one when it comes to crossing infrastructure for the convenience of pedestrians. Therefore, crosswalk locations are requested using the tag `highway=crossing`. If the request is successful, the responses carry the crosswalk locations of the regions of interest.

---
[1] http://www.openstreetmap.org
[2] http://wiki.openstreetmap.org/wiki/OverpassAPI

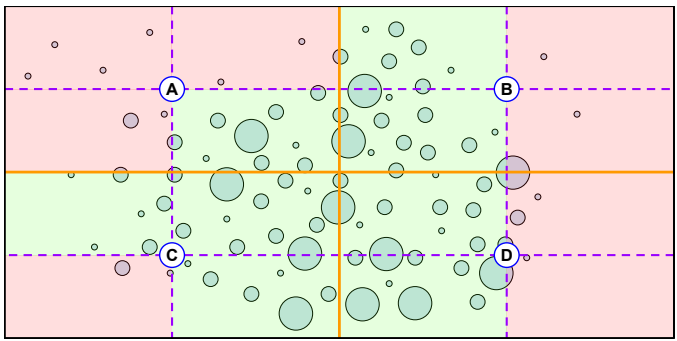

Figure 2: Regions and sub-regions definition. A region (black solid line) must be compliant with Overpass API limitations. If any dimension is larger than 1/4 degrees, the region is split into sub-regions (A, B, C and D – orange solid lines) that comply the limitation. If any of these sub-regions contain between 50 and 2000 crosswalks (green sub-regions), the sub-region is kept. Otherwise, if it contains less than 50 crosswalks (red sub-region), it is discarded. If it contains more than 2000, the split is applied recursively, until all regions have been either discarded or kept. Blue circles indicates crosswalk density, i.e. larger the circle, more crosswalks in that area.

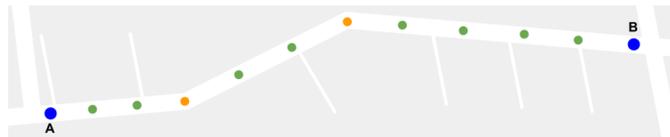

Figure 3: Location Augmentation. Given two crosswalk locations (A and B), a path is requested to the Google Maps Directions API. The path delivered by the API comprises a list of points (orange points, besides the origin and destination) encoded into a polyline. Finally, this list of locations is augmented by evenly sampling locations (green circles) between the points.

With the crosswalk locations in hand, the first strategy to avoid noise is performed. This strategy comprises splitting each sub-region that contains more than 2000 crosswalks into multiple other sub-regions with equal dimensions. This split follows the same procedure performed on the regions. If the resulting sub-region contains less than 50 crosswalks (red sub-regions in the Figure 2), it is discarded. Otherwise, if it contains less than 2000 and more than 50 crosswalks (green sub-regions in the Figure 2), it is kept. If the sub-region still have more than 2000 crosswalks, the process is applied recursively until all sub-regions have more than 50 and less than 2000 crosswalks. These values were empirically defined. This process, illustrated in the Figure 2, attempts to remove areas of the regions of interest that present low density of crosswalk annotations. These areas may have been inconsistently annotated or they simply contain fewer crosswalks. In both cases, these areas are likely to contain crosswalks that were not annotated, which might result in false negative samples.

*3.1.2. Locations Retrieval*

At this point, all crosswalk locations of all regions of interest were retrieved. Nevertheless, neither positive nor negative samples can be acquired yet. From the perspective of a car or a driver (cockpit view), images that contain crosswalks are taken from locations near to crosswalks, i.e. not at the crosswalk locations themselves. In other words, there is still need to discover places nearby crosswalk locations. The location of these nearby places are discovered using the following procedure. Initially, a path between two known crosswalk locations (both randomly chosen within a sub-region) is requested to the Google Maps Directions API. To reduce the number of requests to this API, each request contains up to 23 subsequent crosswalk locations (blue circles in the Figure 3), besides the origin and destination. The Google Maps Directions API returns by default the directions in the driving mode, i.e. that a car could follow to go from one place to the other. These directions are returned as encoded polylines. The system decodes these polylines into a list of points (orange circles in the Figure 3), i.e. locations. However, these points are distributed in order to optimize the encoding process, resulting in points that are not evenly distributed.

Positive samples are necessarily near to the crosswalk locations, therefore the list of locations derived from the polylines needs to be augmented in order to generate more samples to be collected near by the crosswalk locations. The augmentation (see Figure 3) is performed by sampling locations between two subsequent points at equal distances. The sampling strategy ensures that each sampled location (green circles in the Figure 3) is at most $1 \times 10^{-4}$ degrees ($\approx 11$ meters) from another location. At the end of the augmentation process, the proposed system has a much larger list of locations that can be used to retrieve both positive and negative samples, and capture images with larger diversity.

Finally, this list needs to be cleared of duplicate locations. Duplicate locations may appear because the paths provided by the Google Maps Directions API may be partially shared across different requests, i.e. there may be partial path overlap between two or more path requests. Besides that, the augmentation process can also generate points in locations already existent in another paths.

*3.1.3. Samples Acquisition and Automatic Annotation*

To request positive and negative samples (images) to the Google Static Maps API[3], there are some parameters to be set. Two parameters are mandatory: location and size. The location (in world coordinates, i.e. latitude and longitude) varies according to the position of the sample, and the size is fixed as 640×520 pixels. Besides these mandatory parameters, there are three others that influence the resulting image and are optional: field of view, pitch and heading. In the first two parameters (field of view and pitch), the default values were used: 90 and 0 degrees, respectively. As the documentation says, the field of view (fov) can be seen as a zoom factor, i.e. all the images will present the same "zoom factor", and the pitch is the vertical angle of the camera relative to the vehicle, i.e. the default angle is often flat horizontal. The last one (heading) is basically the direction that the camera is pointed to. Differently than the previous ones, the direction needs to be carefully calculated in order to retrieve useful images.

---
[3] http://developers.google.com/maps/documentation/static-maps

Before defining the direction for each location, an additional step is performed. This additional step is required because the Street View images come from specific locations, i.e. the locations defined so far are likely not to be the exact locations of available Street View images. To retrieve the location of the nearest Street View image of a given location, the proposed system needs to dispatch a request to the Street View Image Metadata API. This API requires a location and returns the latitude and longitude of the nearest Street View panorama to the given location. Besides that, the id of the panorama, the date the photo was taken, and copyright information are also provided by the Street View Image Metadata API. If there is no Street View image available near (i.e. in a radius of 50 meters) to the requested location, the requested location is discarded by our system. As a result of this step, the input list of locations is turned into a list of valid Street View locations. There are three important notes about this step. First, if this step is not performed, the direction for each location can be wrongly calculated, and the resulting image may differ from the expected. Second, two different locations in the input list may be represented by the same Street View image location, and this may lead to duplicate locations. For that reason, only unique locations remain in the resulting list of valid Street View locations. Third, when the requests of the proposed system were dispatched, there was a bug that caused requests to Street View Image Metadata API to consume the quota, and this quota is shared with the Google Street View Image API (used to request the images). Therefore, the least requests, the better. However, this bug has been fixed by Google[4].

After the generation of the list of valid Street View image locations, the proposed system is ready to calculate the heading for each location and request the images. The heading ($\alpha$) for each location is defined by the Equation 1.

$$\alpha = \text{atan2}(y_{i+1} - y_i, x_{i+1} - x_i) \cdot \frac{180}{\pi} \quad (1)$$

where $\alpha$ is defined in degrees, and $x_i$ and $y_i$ are the latitude and longitude of the i-th valid Street View location (e.g. A' and C' in the Figure 4), respectively. After the definition of the heading for every location, the system is ready to dispatch the requests of both positive and negative samples to the Google Street View Image API.

Finally, the proposed system has all the images and it needs to properly annotate every single of them. Basically, the samples are separated between positive and negative using a simple rule (illustrated in the Figure 4). The image is only considered as positive sample if a crosswalk location is within the field of view of the image and in a certain range. The proposed system uses a narrower field of view (70°, from $\alpha - 35°$ to $\alpha + 35°$) than the one requested to the API (90°), i.e. it intends to reduce false positives by being more restrictive. Besides that, a crosswalk must be located farther than $5° \times 10^{-5}$ ($\approx 5.6$ meters) and closer than $2.5° \times 10^{-4}$ ($\approx 27.8$ meters). If any crosswalk is located within this region (green region in the Figure 4), the image is annotated as a positive sample. Otherwise, it will be labeled as a negative sample.

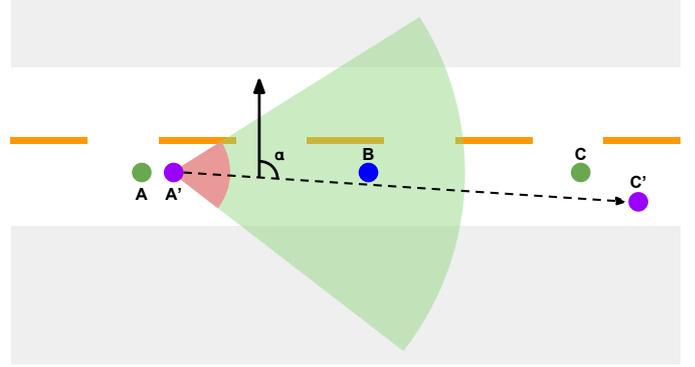

Figure 4: Sample Acquisition and Annotation. Given three locations (A, B and C), where only B is a crosswalk (blue circle), valid Street View panorama locations are requested to the Street View Image Metadata API. The nearest panorama locations (A' and C') are represented by the purple circles. The heading of a location is defined by $\alpha$. If a crosswalk location is in the green area, the image from A' is considered as a positive sample. Otherwise, it is a negative sample.

*3.1.4. Partial Annotation*

The crosswalk locations were retrieved using the Overpass API (OpenStreetMap data). The OpenStreetMap is a mapping initiative based on crowdsourcing, i.e. the data provided by the OpenStreetMap is the result of the contributions of their voluntary users. The quality of such data is based on self-regulation. Nevertheless, data provided by voluntary users may not be as accurate as required by our system. Therefore, despite of the efforts to automatically reduce false positive and negative samples, the dataset automatically acquired can be noisy. In this context, the user has the opportunity to manually annotate part of the data in order to increase the final accuracy of the system. To show the increase in accuracy with manual annotation, part of the dataset was re-annotated by a human expert. The amount of negative samples is far greater than the amount of positive samples, as observed in the real world. Because of that, only the crosswalks were re-annotated, thence are refereed to as partial annotation. The annotations were made by a human which used two subjective criteria: *i)* the crosswalk should be occupying a reasonable area of the image, i.e. crosswalks too far should be removed (see Figure 5 (a) and (b) for positive and negative cases using this criterion); *ii)* the crosswalk in the image should be preferably in the traffic direction of the vehicle (see Figure 5 (c) and (d) for positive and negative cases using this criterion). In total, 17.36% of the 27,241 manually annotated images had their labels changed, i.e., the human expert considered them as wrongly annotated by the automatic procedure. The majority of these samples erroneously annotated were automatically annotated as positive samples: 38.24% changed from negative to positive and 61.76% from positive to negative.

---
[4] https://issuetracker.google.com/issues/35830093

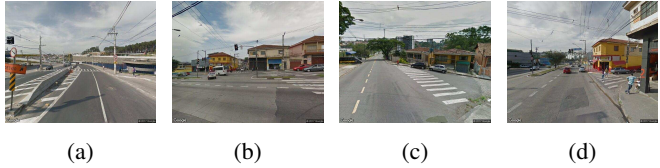

| (a) | (b) | (c) | (d) |

Figure 5: Sample images manually annotated during the partial annotation. The images (a) and (c) were manually labeled as positive samples. The sample (b) was labeled as negative because the crosswalk occupies a very small area in the image, i.e. it is too far. The sample (d) was labeled as negative because the crosswalk is not in the traffic direction of the vehicle.

### 3.2. Convolutional Neural Network

In the proposed system, the dataset acquired is used to train a Convolutional Neural Network (ConvNet). The system uses the VGG architecture. This architecture was proposed by Simonyan and Zisserman [12] to the ImageNet Large Scale Visual Recognition Challenge (ILSVRC 2012) [13]. The net contains 16 layers: 13 convolutional followed by 3 fully connected layers. The last fully connected layer had 1000 neurons because of the 1000 categories of the ILSVRC. For that reason, the last layer was changed to only 2 neurons, one for each class of the crosswalk classification problem: crosswalk (positive), i.e. the image contains a crosswalk; and no-crosswalk (negative), i.e. the image does not contain a crosswalk.

### 3.3. Model Training

For the training of the model, all the images automatically acquired were used as input. The model expects an input of size $256 \times 256$ pixels. For that reason, all images were rescaled from their original size ($640 \times 520$) down to $256 \times 256$ using bilinear interpolation. Moreover, a data augmentation procedure is also performed. The data augmentation is performed on-the-fly, and comprises two operations: *i)* cropping $224 \times 224$ pixels, *ii)* mirroring horizontally the image. Both operations are performed randomly, i.e. a random position is chosen to the crop, and there is a 50% chance of mirroring the image. The result of this data augmentation is provided as input to the ConvNet. During the training, the network weights were initialized randomly using the Xavier [14] algorithm. Although fine-tuning procedures are widely used in the literature, preliminary analysis showed they achieved: i) results on pair with the randomly initializing the weights, i.e., there was no significant statistical difference; ii) no significant improvement in the learning time, i.e., the convergence rate was only slightly improved; and iii) worse results when they were evaluated in the cross-database (especially in the IARA dataset) compared to the ones with randomly initialized weights. Therefore, fine-tuning is not employed in the models hereby presented. All the models were trained during 10 epochs using an initial learning rate of $1\times10^{-2}$ and a Step Down policy that decreases the learning rate by a factor of 10 three times during the training, i.e. at the end of the training process, the learning rate was equal to $1 \times 10^{-4}$. After the 10 epochs, the model is considered trained. Two models were trained: GSV-FA and GSV-PA. The former was trained using the fully-automatic (hence FA) dataset, i.e. all the images were automatically annotated by the system. The latter was trained using the partially-automatic (hence PA) dataset, i.e. all the crosswalks were manually re-annotated by an expert. In both cases, the system used all the crosswalks available and two times the amount of positive samples as negative samples. The negative samples were randomly chosen.

### 3.4. Model Inference

At this point, the dataset was automatically acquired and annotated, and a model is already trained. At inference, images from different image sources may be used. Therefore, all input images are downsampled to $256 \times 256$ pixels using bilinear interpolation. After that, because of the data augmentation procedure performed during the training, a centralized crop of $224 \times 224$ pixels has to be performed. This cropped image is forwarded through the ConvNet. Finally, the ConvNet yields a probability for each class: crosswalk (positive) and no-crosswalk (negative).

## 4. Experimental Methodology

In this section, the methodology and materials used in the experimentation process are described. Firstly, the datasets used for the training and evaluation of the system are detailed. Secondly, the metrics used for the quantitative experimentation are presented. Finally, the experiments are described in details.

### 4.1. Datasets

Three datasets from different sources were used during the training and evaluation of the proposed system: the dataset from the Google Street View including the automatically acquired and annotated versions; one acquired by the camera system of an autonomous car; and, one acquired by a camera placed inside a standard car. These datasets are also described below. The scripts used for the acquisition of the Google Street View dataset will be made publicly available. In addition, the image sequences of the other two datasets (more than 23,500 images) will also be publicly released.

#### 4.1.1. Google Street View – GSV

The Google Street View (GSV) dataset was automatically acquired and annotated by the proposed system. The images of this dataset came from Google StreetView (via Google Static Maps API) using the procedure described previously in the Section 3. It comprises non-sequential images of two classes: crosswalks and no-crosswalks. All the images (both positive and negative) are from Brazil. Brazilian roads present all sorts of challenges expected to be encountered in the roads worldwide. Crosswalks alone can be presented in a variety of ways: the painting can be fading away, aging; there can be pedestrians, vehicles, and other objects partially occluding the crosswalks; shadows of building, trees, large vehicles and other objects may be partially or completely darkening the road; crosswalks themselves have different styles (e.g. background colored in red, blue; with arrows on it; etc.). All these challenges can be seen in this dataset. Because of the process of retrieving locations used by the system, most images are within a road heading in the traffic direction. The size of the images was defined to 640 × 520. The default values of the API for both field of view and pitch were used: 90 and 0 degrees, respectively. Their documentation states that the default pitch is often, but not always, flat horizontal.

This dataset comprises 657,691 images from 20 states of the Brazil plus the federal district (Brazil has 26 states, in total). In total, there are 33,959 images of crosswalks (i.e. positive samples) and 623,732 images of no-crosswalks (i.e. negative samples). The dataset is divided into 24 regions. Even though each region is named after a city, regions may include nearby towns besides the city they are named after. The cities chosen are mainly capitals and big cities. Many factors contributed to this decision: the images on the Google Street View in the capitals are more likely to be recent and capitals are also more likely to contain crosswalk annotations in the OpenStreetMap. Some samples of the GSV dataset can be seen in the Figure 6.

In order to properly evaluate on this dataset, part of the database was separated as test set. To build the test set, some regions were chosen to provide the images. These regions were not in the training set, therefore ensuring that images in the test set are not in the training set. The test set was entirely manually labeled (27,241 images in total). Firstly, all positive samples were manually labeled following the same criteria used on the partial annotation of the dataset. Secondly, the number of negative samples to be annotated was defined as twice the number of manually annotated samples (a typical proportion of real-world use case [2]). Finally, negative samples were annotated until the amount of truly negative samples reach twice the number of truly positive samples. This fully-manual annotated set is refereed as test set, and it was used in the evaluation of the GSV-FA and GSV-PA models on the first experiment (Evaluation on the GSV dataset). This proportion (2:1) between the negative and positive samples was also ensured in the training and validation sets. The training and validation sets comprise all positive samples of the training regions plus randomly sampled negative ones (ensuring the 2:1 factor). The dataset splits (training, validation, and test sets) used in the experimentation are detailed in the Subsection 4.4.1.

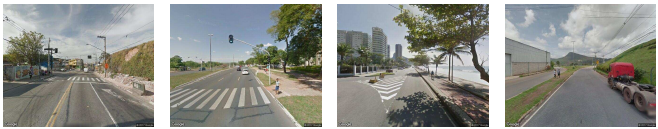

Figure 6: Sample images of the GSV dataset. First two images are positive samples (contain crosswalks) and the last two are negative samples.

### 4.1.2. IARA

In order to assess the generalization of the trained model, the proposed system was evaluated in datasets from different sources. The IARA dataset is named after the vehicle used to capture the images: the Intelligent Autonomous Robotic Automobile. IARA is an autonomous vehicle that is being developed in the High Performance Computing Lab (LCAD) of the Universidade Federal do Espírito Santo. The vehicle contains many sensors, but only one camera was used to record this dataset. The camera, a Bumblebee XB3, was mounted on the top of the car facing forward. The images generated by this camera are very different from the ones in the GSV dataset, used during training. Therefore, the evaluation on this dataset is challenging. This dataset was recorded during the day in a week-day, i.e. it contains usual traffic. In total, there are 12,441 images from 4 different sequences recorded in the capital of the Espírito Santo, Vitória. Differently from the GSV dataset, the IARA dataset had to be manually labeled. An image was annotated as positive sample following the same criteria used in the partial annotation of the GSV dataset. In total, 2,637 images (21.2%) were labeled as positive samples and 9,804 images (78.8%) as negative samples in the IARA dataset. Some samples of the IARA dataset can be seen in the Figure 7.

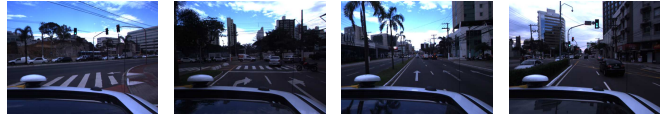

Figure 7: Sample images of the IARA dataset. First two images are positive samples (contain crosswalks) and the last two are negative samples.

### 4.1.3. GOPRO

The GOPRO dataset comprises 11,070 images recorded in the city of Vitória, Vila Velha and Guarapari, Espírito Santo, Brazil. The videos were recorded using a GoPRO HERO 3 camera in Full HD (1920 × 1080 pixels) at 29.97 frames per second (FPS) in different days. Some of them were recorded in city roads and the others in the highways connecting these cities. The images are divided into 29 sequences. From them, 23 of them are short sequences (up to 15 seconds) of a vehicle passing by crosswalks and the other 6 are longer sequences (up to 90 seconds) of a vehicle driving without any crosswalk in the field of view. All the videos were manually annotated in order to enable both qualitative and quantitative evaluation. The annotation followed the same criteria used on the partial annotation of the GSV dataset. In total, crosswalks were manually annotated in 17.34% of the images. It is worth noting that the process of annotating a crosswalk in a sequence is subjective, i.e. when a vehicle is approaching a crosswalk, it is really hard to tell when the annotation of that crosswalk should begin. For that reason, qualitative evaluation is essential to support the quantitative results. The crosswalks in these sequences are presented in a variety of ways, such as with pedestrians, occluded by cars, painting fading away (i.e. aging), etc. Some images of both positive and negative samples are shown in the Figure 8.

### 4.2. Metrics

For the evaluation of the proposed system, two metrics were used: global accuracy and $F_1$ score. The global accuracy is defined as in the Equation 2, and the $F_1$ score definition can be seen in the Equation 3:

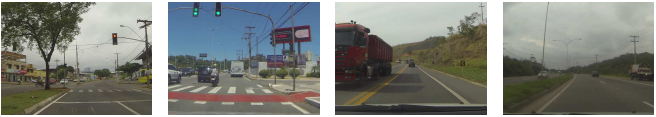

Figure 8: Sample images of the GOPRO dataset. First two images are positive samples (contain crosswalks) and the last two are negative samples.

$$\text{ACC} = \frac{\text{TP} + \text{TN}}{\text{P} + \text{N}} \quad (2)$$

$$F_1 = 2 \cdot \frac{\text{precision} \cdot \text{recall}}{\text{precision} + \text{recall}} \quad (3)$$

where TP, TN, P and N means the number of True Positive, True Negative, Positive and Negative, respectively. Precision is defined by $\frac{\text{TP}}{\text{TP+FP}}$ and recall by $\frac{\text{TP}}{\text{TP+TN}}$, where FP means False Positive and FN means False Negative.

Besides the accuracy and $F_1$ score, another metric is proposed, because these two metrics (ACC and $F_1$) are image-based metrics, i.e. they do not consider the temporal dimension of a dataset. In order to cope with that, an instance-based metric is proposed. This metric ($\text{ACC}_{INSTANCE}$) considers crosswalks as instances. Therefore, a hit is considered only if the proposed system correctly classified the crosswalk at least half of its time in the sequence. In other words, if a crosswalk is shown in 20 sequential images (frames) of a sequence, the crosswalk classification will only be considered as correct if the proposed system correctly classified it for more than 10 of these frames. Therefore, the $\text{ACC}_{INSTANCE}$ reports how many of the crosswalks the proposed system correctly identified in a given temporal sequence of images (e.g., a video). This metric was only used in the videos datasets, the IARA and the GOPRO, because they are temporal sequences of images.

*4.3. Statistical Analysis*

To properly compare the proposed models (GSV-FA and GSV-PA), a pairwise statistical comparison was carried out. For the statistical analysis, we have used the paired *t*-test [15]. In this test, low p-values mean that the distributions are indeed significantly different. As a threshold, if the p-value is lower than $1 \times 10^{-4}$, the difference can be considered significant.

*4.4. Experiments*

Some experiments were designed for the evaluation of the proposed system. These experiments intend to assess the generalization of the system and to measure its performance in terms of the proposed metrics. The quantitative experimentation can be unfolded in two experiments, one for each dataset. In the first experiment, the models were trained on several different combinations of the GSV-FA and GSV-PA datasets and evaluated on the GSV test set. Then, the best models of the first experiment are used on the cross-database experiments: the model trained on the GSV dataset is evaluated on the IARA and GOPRO datasets. Besides that, qualitative experimentation was also performed. These experiments are detailed below.

*4.4.1. Quantitative*

Quantitative experiments intend to report the accuracy and the $F_1$ score for the evaluation on the GSV dataset, and all three metrics (ACC, $F_1$, and $\text{ACC}_{INSTANCE}$) for the cross-database experiments.

*Evaluation on the GSV dataset.* In the first part, the proposed system is used to train two models: one using the fully automatically annotated images (GSV-FA) and another using the partially automatically annotated images (GSV-PA), where FA and PA stands for fully-automatic and partially-automatic, respectively. The difference between them is that in the former all the images are labeled according to the automatic procedure of the proposed system, and in the latter only the crosswalks images (i.e. positive samples) were manually labeled. This experiment intends to measure the advantage of having part of the data manually annotated. The dataset splits used for the training was carefully chosen to ensure a fairer evaluation protocol. First, the GSV dataset was split into three sets: train, validation and test. The test set was chosen region-wise, i.e. none of the regions in the test set was seen during training or validation. The test regions were randomly chosen trying to keep approximately 25% of the database in the test set. The train and validation splits comprise the rest of the data, where 90% of it was used to train the models and only 10% for the validation, and together they comprise 17 regions. Second, during the training, the number of negative samples was equivalent to $\approx 2$ times the number of positive samples in both models. The negative samples were randomly sampled, but as the test set comprises mutually exclusive regions, none of the negative samples used in any of the training sets were used in the test set. This imbalance was used to approximate real-world use cases, where most of the time there are no crosswalks in the field of view of the driver. As a result, each fully automatic training set contains 65,095 images and each validation set contains less than 7,000 images. Finally, in order to enable a fairer pairwise comparison of the experiments, the GSV-FA dataset was generated based on the GSV-PA, i.e. GSV-PA is always a subset of GSV-FA. As the number of crosswalks in the GSV-FA is greater than in the GSV-PA, several other no-crosswalks images (complementary set) were randomly chosen to create GSV-FA. This complementary set comprises non-duplicate images from the training/validation split. Each training set of the GSV-PA models contains 39,302 images and each validation set contains less than 4,500 images.

*Cross-Database.* The IARA and GOPRO datasets were used for the cross-database experiments. For these experiments, the best models trained in the evaluation on the GSV dataset are used to be evaluated in the other datasets. These models are referenced as GSV-FA* and GSV-PA* (the best model trained on fully-automatic and partially-automatic dataset, respectively). This experiment was planned to evaluate the generalization of the model. In fact, the models hereby evaluated are trained on images from Google Street View and evaluated on real-world scenarios from different sources. It was ensured that the regions of the IARA and GOPRO datasets were not in the training set of the evaluated models. As these datasets comprise sequential images, the $ACC_{INSTANCE}$ is also reported.

*4.4.2. Qualitative*

In addition to the quantitative experiments, the cross-database experiments (IARA and GOPRO datasets) were recorded and are available as supplementary to allow a qualitative evaluation. As already discussed, the qualitative evaluation is essential to support the quantitative evaluation, specially on sequences of images where the decision on when a crosswalk begin is fairly subjective. In total, more than 23,500 images (more than 30 sequences) were used on the qualitative evaluation.

*4.5. Setup*

The experiments were carried out in an Intel Core i7-4770 3.40 GHz with 16GB of RAM, and 1 Tesla K40 GPU with 12GB of memory. The machine was running Linux Ubuntu 14.04 with NVIDIA CUDA 7.5 and cuDNN 5.1 [16] installed. The training and inference steps were done using a NVIDIA fork of the Caffe framework [17]. In this setup, the training sessions took 4 hours, on average; and the inference can be done in more than 30 frames per second. The code for the dataset acquisition and the experiments were written in Python, using the OpenCV library. The source-code of the proposed system and the experiments performed in this work will be made publicly available[5].

## 5. Results and Discussions

The results of the proposed experiments are presented in this section. At first, the results of both models (GSV-FA and GSV-PA) in the evaluation on the GSV dataset are shown to identify the models with the higher overall accuracy. After that, the results of these models (GSV-FA* and GSV-PA*) on the cross-database experiments are presented. Qualitative results are shown for all the experiments.

*5.1. Results of Evaluation on the GSV dataset*

In this experiment, two models are evaluated. Initially, the results of the model trained on the fully-automatic dataset (GSV-FA) is presented. The results of the 10 runs (each run with a different negative subset) can be seen in the Table 1. On average, the model achieves an overall accuracy of 94.12%, and 89.00% of $F_1$ score.

As can be seen in the Table 1, the GSV-FA models achieved high performance results. Besides these results, the very low standard deviation (0.21%) indicates that the models are very robust to the noise in the dataset, specially to the many variations of no-crosswalk subsets presented to it, and to the initialization of the weights. Moreover, Table 1 shows the GSV-FA model presents a balanced accuracy between the positive and negative classes.

[5]https://github.com/rodrigoberriel/streetview-crosswalk-classification

Table 1: Accuracy and $F_1$ score of the Model Trained on the Fully Automatic Dataset (GSV-FA)

| # | Accuracy (%) | | | $F_1$ score (%) |
|---|---|---|---|---|
| | **Overall** | **Negative** | **Positive** | |
| 1 | 93.85 | 93.21 | 95.76 | 88.67 |
| 2 | 94.21 | 94.02 | 94.75 | 89.15 |
| 3 | **94.38** | 94.27 | 94.70 | 89.43 |
| 4 | 94.15 | 94.06 | 94.43 | 89.03 |
| 5 | 94.22 | 93.97 | 94.96 | 89.19 |
| 6 | 94.15 | 93.91 | 94.89 | 89.08 |
| 7 | 94.27 | 94.67 | 93.09 | 89.09 |
| 8 | 93.74 | 93.43 | 94.67 | 88.38 |
| 9 | 94.32 | 94.24 | 94.56 | 89.32 |
| 10 | 93.95 | 93.81 | 94.39 | 88.69 |
| **Average** | **94.12** ± **0.21** | **93.96** ± **0.42** | **94.62** ± **0.66** | **89.00** ± **0.33** |

After evaluating the fully-automatic model, the results of model trained on the partially-automatic dataset (GSV-PA) are shown. The results of the 10 runs can be seen in the Table 2. For each run, the variations are generated by the different subsets of negative samples and the random initialization of the weights. On average, the GSV-PA model achieves 96.30% of overall accuracy and 92.78% of $F_1$ score.

Table 2: Accuracy and $F_1$ score of the Model Trained on the Partially Automatic Dataset (GSV-PA)

| # | Accuracy (%) | | | $F_1$ score (%) |
|---|---|---|---|---|
| | **Overall** | **Negative** | **Positive** | |
| 1 | 96.42 | 97.18 | 94.15 | 92.97 |
| 2 | 96.31 | 96.90 | 94.58 | 92.80 |
| 3 | 96.41 | 97.22 | 94.01 | 92.94 |
| 4 | 96.36 | 96.88 | 94.83 | 92.91 |
| 5 | **96.51** | 97.14 | 94.62 | 93.16 |
| 6 | 96.06 | 96.43 | 94.94 | 92.37 |
| 7 | 96.26 | 96.94 | 94.21 | 92.68 |
| 8 | 96.18 | 96.59 | 94.94 | 92.58 |
| 9 | 96.14 | 96.53 | 94.97 | 92.51 |
| 10 | 96.39 | 97.16 | 94.08 | 92.90 |
| **Average** | **96.30** ± **0.14** | **96.90** ± **0.29** | **94.53** ± **0.39** | **92.78** ± **0.24** |

The Table 2 shows that the GSV-PA models were also able to achieve very high performance results. The low standard deviation (0.14%) indicates that these models were also robust to the training variations.

In addition to the experiments with the VGG network, we performed experiments with the AlexNet [18] architecture. The results of the AlexNet models are on pair with the VGG network, which demonstrates it is possible to use a smaller network (e.g., in GPUs with lower memory) without losing much performance. We also performed an experiment on the GSV-FA with a simpler learning method (Multilayer Perceptron – MLP) by naively plugging it in the input image (since hand-crafted features are not within the scope of this study). The MLP models yielded much worse results when compared to the ConvNets (both VGG and AlexNet). It shows the benefits of a representation learning (end-to-end classification) approach.

We further explored the errors of the models, both positive and negative. We noticed that most of the positive errors (i.e., manually labeled as positive and predicted as negative) presented mainly four types of errors: *i)* genuine mistakes; *ii)* strong occlusions; *iii)* aging crosswalks (i.e., painting fading away); and *iv)* different types of crosswalks (i.e., types that are naturally less frequent such as with blue and red backgrounds, dashed style, etc.). In addition, most of the negative errors (i.e., manually labeled as negative and predicted as positive) also presented four types of errors, in general: *i)* genuine mistakes; *ii)* crosswalks that are too far; *iii)* arrows, writings, and other road markings; and *iv)* crosswalks not in the traffic direction. As can be seen in the videos of the qualitative analysis, the models predict correctly many samples in these situations (e.g., road markings, crosswalks with red background, etc.), but these are the ones they are more likely to fail.

In comparison, the GSV-PA models presented an improvement of the overall accuracy from 94.12% to 96.30% (+2.18%), as expected. Furthermore, the high performance results presented by both partially-automatic (GSV-PA) and fully-automatic (GSV-FA) models show that the GSV-FA models are very robust to the noise in the dataset. The difference is relatively small when the amount of human effort is accounted. Despite of it, this effort may be worth depending on the application. Statistical analysis shows that the difference between these two models (GSV-FA and GSV-PA) are indeed significant, i.e. p-value $\approx 5.5 \times 10^{-10}$ for the overall accuracy and p-value $\approx 2.3 \times 10^{-10}$ for the $F_1$ score.

### 5.2. Results of the Cross-Database Experiment

In the cross-database experiment, the best models (GSV-FA* and GSV-PA*) chosen in the evaluation on the GSV dataset are evaluated in other datasets from different sources (e.g., different cameras). The first of the cross-database experiments was carried out on the IARA dataset. The GSV-FA* model achieved 95.65% of ACC$_{INSTANCE}$. Evaluating frame-by-frame, the GSV-FA* model achieved an overall accuracy of 95.46% ($F_1$ score of 90.03%). The GSV-PA* model achieved an overall accuracy of 90.20% ($F_1$ score of 75.61%), which is slightly worse than the GSV-FA*. In addition, the GSV-PA* achieved 89.13% considering the ACC$_{INSTANCE}$. An example of the performance of the GSV-FA* and GSV-PA* models on the IARA dataset can be seen in the Figure 9. It is important to note that GSV-PA* reported lower $F_1$ score when compared to the GSV-FA* mainly because of two long sequences where the vehicle was stopped in a red traffic light with vehicles highly occluding the crosswalk. These sequences were annotated as positive samples, despite of the high occlusion. On the other hand, surprisingly, the GSV-FA* was able to predict these highly occluded crosswalks consistently.

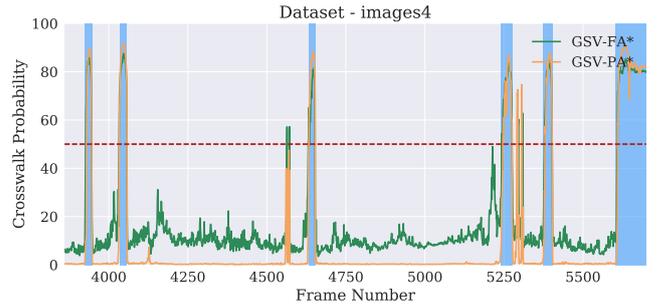

Figure 9: Performance of the GSV-FA* and GSV-PA* models on a sequence of the IARA dataset. The red dashed line at 50% represents the threshold for the prediction: if it is below the threshold, the model is predicting no-crosswalk. Otherwise, the model predicts that there is a crosswalk in the image. Blue regions are those manually labeled as having crosswalks. The failure (false positive) case near the frame #4500 is the first image of the Figure 10.

As can be seen in the Figure 9, the GSV-PA* model is fairly more robust in the negative predictions. When it comes to the positive predictions, the GSV-PA* predicts correctly with an average of 90.30% of confidence compared to 91.88% of the GSV-FA*. Additionally, failure cases on the IARA dataset are shown in the Figure 10.

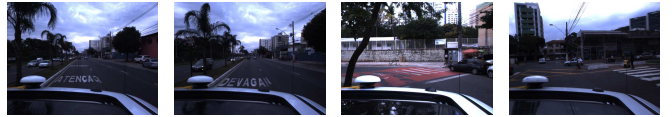

Figure 10: Failure samples on the IARA dataset. The first two images are false positives, and the last two are false negative predictions.

For the second database of the cross-database experiments, the models were evaluated on the GOPRO dataset. On this dataset, the GSV-FA* model achieved an overall accuracy of 88.16%, and 92.31% on the ACC$_{INSTANCE}$ ($F_1$ score of 72.01%). As expected, the GSV-PA* model achieved slightly better results: overall accuracy of 92.85% and ACC$_{INSTANCE}$ of 96.15% ($F_1$ score of 82.07%). An example of the performance of the GSV-FA* and GSV-PA* models on the GOPRO dataset can be seen in the Figure 11.

The GSV-PA* model achieved better results on both the overall accuracy (frame-based metric) and on the ACC$_{INSTANCE}$ (crosswalk-based metric). This can be explained by the fact that the GSV-FA* model is less stable than the GSV-PA*: during the crosswalks, the GSV-PA* predicted crosswalk for 82.28% of the samples, on average, against 88.56% of the GSV-FA*. Additionally, failure cases on the GOPRO dataset are shown in the Figure 12.

As can be seen in the Table 3, the models trained using the proposed system can indeed achieve good performance results even on datasets from different sources, i.e. images with clear differences on the brightness, contrast, size, etc. Based on the results, the IARA dataset seems more difficult than the GOPRO dataset. Looking at the images, the camera used on the IARA dataset seems to be calibrated very differently than the other datasets. This difference may have lead to unexpected inconsis-tencies, such as the GSV-FA* achieving better results than the GSV-PA* for the IARA dataset.

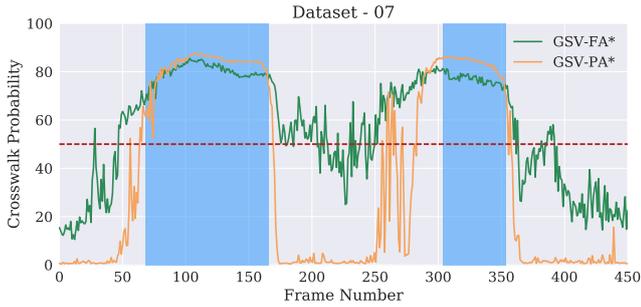

Figure 11: Performance of the GSV-FA* and GSV-PA* models on a sequence of the GOPRO dataset. The red dashed line at 50% represents the same threshold as in the Figure 9. Between the frames #250 and #300 both models begin to report a crosswalk. However, the human annotator only labeled it after the frame #300. This illustrates the importance of the qualitative evaluation to support the quantitative results. The frame #275 of this dataset can be seen in the first image of the Figure 12.

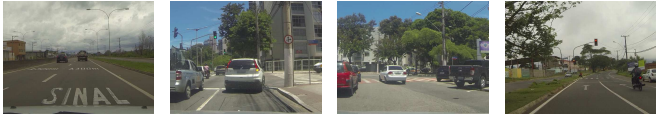

Figure 12: Failure samples on the GOPRO dataset. The first two images are false positives, and the last two are false negative predictions. The second image is particularly difficult, because there is a crosswalk in the intersection and it does not fit into the criteria used for the manual annotation.

Table 3: Performance overview of the best fully-automatic (FA) and partially-automatic (PA) models using both intra-database (GSV) and cross-database (DATASET-X and GOPRO) evaluation protocols.

| Model   | Dataset | ACC    | ACC$_{INSTANCE}$ |
|---------|---------|--------|------------------|
|         | GSV     | 94.21% | –                |
| GSV-FA* | IARA    | 93.06% | 81.25%           |
|         | GOPRO   | 89.30% | 92.31%           |
|         | GSV     | 96.51% | –                |
| GSV-PA* | IARA    | 92.04% | 89.58%           |
|         | GOPRO   | 93.48% | 96.15%           |

Comparing the results hereby presented to the literature is very difficult. Most of the works in the literature do not release publicly the dataset used in their experimentation nor their methods, therefore it is hardly possible to replicate their results. Even though, given the appropriate considerations, some results of the literature can provide context to ours. Wang et al. [3] proposed an SVM classifier and achieved an overall accuracy of 90.98%, and 78.90% for the positive class. The evaluation of their method was performed in a small (122 images, in total) and local dataset. Riveiro et al. [4] proposed the combination of several image processing techniques to detect crosswalks. They evaluated their method in a small (30 crosswalks) and local dataset, and achieved 83.33% of accuracy for the positive class. Poggi et al. [9] proposed the use of CNNs to detect crosswalks in the perspective of a person. They proposed two versions: without any refinement (88.97%) and with head pose refinement (91.59%). These results were achieved evaluating both of their models in a local and smaller (10,165 frames) dataset compared to ours. As discussed in the section 2, most of the results in the literature rely on an evaluation made on small and local datasets. On the other hand, our models were evaluated in large-scale databases and still presented higher performance results (in absolute terms) than the literature. In addition, none of the methods in the literature performed cross-database evaluation. Nonetheless, our cross-database results also presented performance results on pair with the other experiments in the literature. It is worth noticing that both our databases (IARA and GOPRO) and the scripts to automatically acquire the GSV dataset will be publicly released to enable future fair comparisons.

*5.3. Results of the Qualitative Experiments*

Qualitative results can be viewed on the publicly available videos: using the IARA[6] and the GOPRO[7] datasets. The videos, in total, contain the evaluation on more than 23,500 images from these two datasets. As can be seen in the videos, the model presents robustness to various factors. In addition, the failure cases are consistent, i.e. they tend to happen in some specific situations and not randomly, which indicates the need for more images of those cases during training. The two most common failure cases are when the crosswalk is too far or there are arrows or writings on the road. The first case leads to two failures: false positive, when the human annotator judged that the crosswalk was too far but the model could detect it; and false negative, when the model misses crosswalks that indeed are too far. The problem is that the definition of "too far" is subjective. In those cases, the qualitative analysis may provide useful insights. The second case leads to a much more difficult problem. Currently, there is no way to automatically retrieve images of that particular type (i.e. with arrows or writing in the lane) using the services that were used in this work. Nevertheless, as a future work, we think that we could synthetically augment the dataset by systematically adding arrows and writings to images using a procedure similar to [19]. Note that several post-processing techniques could have been used to increase the performance of the system in those datasets (i.e. temporal sequence of images), such as hysteresis-based technique to remove the few noisy unstable predictions. However, these post-processing techniques were not in the scope of this work. In addition, the same procedure of automatic acquisition and annotation of large sets of images could be extended to other tasks, specially driving related tasks. Some of these extensions may require human-in-the-loop to achieve good performance (e.g., traffic light detection), but others may not. As a future work, we plan to investigate such extensions and the impact of having a human-in-the-loop. In general, qualitative analysis suggests that the models are able to perform very well in real-world scenarios.

---
[6] https://youtu.be/zrKU3duNwuo
[7] https://youtu.be/jmYmQFiqY3c

In addition to the above mentioned experiments, we also performed an experiment with a video recorded during the night using the same camera used in the IARA dataset. The sequence comprises 12,114 frames (in a temporal sequence) covering part of the IARA dataset and more (including a toll, a bridge, etc.) during the night (Figure 13). It is important to note that only daytime images were presented to the models during the training (Google Street View only provides images in daylight). We also ensured that the regions in this night sequence were not in the training sets as well. Although not as stable as in the other datasets, the models were able to correctly predict the existence of the crosswalks in most of the cases, as can be seen in the publicly available video[8].

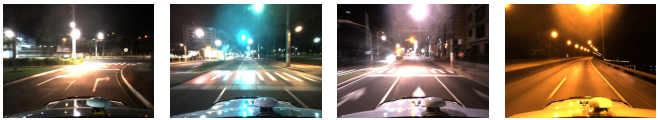

Figure 13: Samples of the dataset used to evaluate the models during the night.

*5.4. Limitations*

The analysis of the qualitative experiments already presented some of the limitations of the proposed system. In addition to those limitations, three others are important to mention. Firstly, the availability of crosswalk annotations in the OpenStreetMap is large but limited, i.e., the system is constrained to acquire positive samples from these locations. Systems such as the one proposed in this work can be used to cope with this fact (helping to annotate other areas) and with the inconsistencies in the data already available (helping to filter the noise). Secondly, the images are acquired from Google Street View, therefore all images are in daylight. Despite of that, experimental results show the model can performed reasonably well in night sequences. Thirdly, given the cost to manually annotate large amounts of images, the manual labels were assigned by a single human annotator, i.e., the subjectivity is constrained to a single individual. To cope with that, subjectivity can be decreased at the high cost of employing a group of annotators. Even though, subjectivity would still apply. Besides that, one application of the models hereby proposed is precisely to avoid this manual labor.

## 6. Conclusion

In this paper, we investigated the use of crowdsourcing platforms for the automatic training of a deep learning model to perform crosswalk classification in Brazil. In addition, we studied the impact of manually annotating a specific part of database on the performance results. Besides evaluating in the database hereby introduced, cross-database experiments were performed.

Our results showed that it is possible to automatically train a deep learning based model via crowdsourcing to accurately classify crosswalks in images. Furthermore, the results also showed that annotating a specific part of the automatically acquired database can boost the performance results. Additionally, the cross-database results showed that the models trained on these automatically acquired datasets can be used to classify crosswalks in images from different sources with high accuracy.

Finally, our results showed that crowdsourcing systems can be exploited to automatically train a crosswalk classification system based on deep learning, and be used on different databases maintaining high accuracy.

**Acknowledgment**

We would like to thank PIBIC/UFES and CAPES for the scholarships, and CNPq for the Grant 311120/2016-4. We gratefully acknowledge the support of NVIDIA Corporation with the donation of the Tesla K40 GPU used for this research. Cloud computing resources were provided by a Microsoft Azure for Research award.

---

[8] https://youtu.be/afCBi1Pj1NE